\documentclass[10pt,twocolumn,letterpaper]{article}

\usepackage{cvpr}              

\usepackage{caption}
\usepackage{algorithm}
\usepackage{algorithmic}
\usepackage{graphicx}
\usepackage{booktabs}
\usepackage{color, soul}
\usepackage{amsmath, amsfonts}
\usepackage{bm}

\usepackage{tikz}
\usepackage{pgfplots}
\usepgfplotslibrary{fillbetween}

\usepackage{array} 
\usepackage{multirow}
\usepackage{enumitem}

\definecolor{cvprblue}{rgb}{0.21,0.49,0.74}
\usepackage[pagebackref,breaklinks,colorlinks,citecolor=cvprblue]{hyperref}

\def\paperID{*****} 
\def\confName{CVPR}
\def\confYear{2024}

\newcounter{alphasect}
\def\alphainsection{0}

\let\oldsection=\section
\def\section{%
  \ifnum\alphainsection=1%
    \addtocounter{alphasect}{1}
  \fi%
\oldsection}%

\renewcommand\thesection{%
  \ifnum\alphainsection=1%
    \Alph{alphasect}%
  \else%
    \arabic{section}%
  \fi%
}%

\newenvironment{alphasection}{%
  \ifnum\alphainsection=1%
    \errhelp={Let other blocks end at the beginning of the next block.}
    \errmessage{Nested Alpha section not allowed}
  \fi%
  \setcounter{alphasect}{0}
  \def\alphainsection{1}
}{%
  \setcounter{alphasect}{0}
  \def\alphainsection{0}
}%

\newcommand{\ourmethod}{TC-PDM~}
\newcommand{\ourmethodnospace}{TC-PDM}

\newcommand{\bbR}{\mathbb{R}}

\newcommand{\cD}{\mathcal{D}}
\newcommand{\bbE}{\mathbb{E}}

\newcommand{\cM}{\mathcal{M}}

\newcommand{\cN}{\mathcal{N}}
\newcommand{\bI}{\mathbf{I}}
\newcommand{\cX}{\mathcal{X}}
\newcommand{\bx}{\mathbf{x}}

\newcommand{\bs}{\mathbf{s}}

\newcommand{\bz}{\mathbf{z}}
\newcommand{\cY}{\mathcal{Y}}
\newcommand{\by}{\mathbf{y}}

\newcommand{\bepsilon}{\bm{\epsilon}}

\title{\ourmethodnospace: Temporally Consistent Patch Diffusion Models for Infrared-to-Visible Video Translation}

\author{
    \vspace{0.5em}
    Anh-Dzung Doan\textsuperscript{\rm 1}, 
    Vu Minh Hieu Phan\textsuperscript{\rm 1}, 
    Surabhi Gupta\textsuperscript{\rm 3}, 
    Markus Wagner\textsuperscript{\rm 2}, 
    Tat-Jun Chin\textsuperscript{\rm 1},
    Ian Reid\textsuperscript{\rm 1} \\ 
     \textsuperscript{\rm 1}Australian Institute for Machine Learning, The University of Adelaide\\
    \textsuperscript{\rm 2}Department of Data Science and Artificial Intelligence, Monash University\\
     \textsuperscript{\rm 3}Safran Electronics and Defense Australasia
}

\begin{document}
\maketitle
\begin{abstract}
Infrared imaging offers resilience against changing lighting conditions by capturing object temperatures.
Yet, in few scenarios, its lack of visual details compared to daytime visible images, poses a significant challenge for human and machine interpretation.
This paper proposes a novel diffusion method, dubbed Temporally Consistent Patch Diffusion Models (TC-DPM), for infrared-to-visible video translation. 
Our method, extending the Patch Diffusion Model, consists of two key components. Firstly, we propose a semantic-guided denoising, leveraging the strong representations of foundational models. As such, our method faithfully preserves the semantic structure of generated visible images. Secondly, we propose a novel temporal blending module to guide the denoising trajectory, ensuring the temporal consistency between consecutive frames.
Experiment shows that \ourmethod outperforms state-of-the-art methods by 35.3\% in FVD for infrared-to-visible video translation and by 6.1\% in AP$_{50}$ for day~$\rightarrow$~night object detection. Our code is publicly available at \url{https://github.com/dzungdoan6/tc-pdm}
\end{abstract}

\section{Introduction}

\begin{figure}
    \centering
    \includegraphics[width=1.0\linewidth]{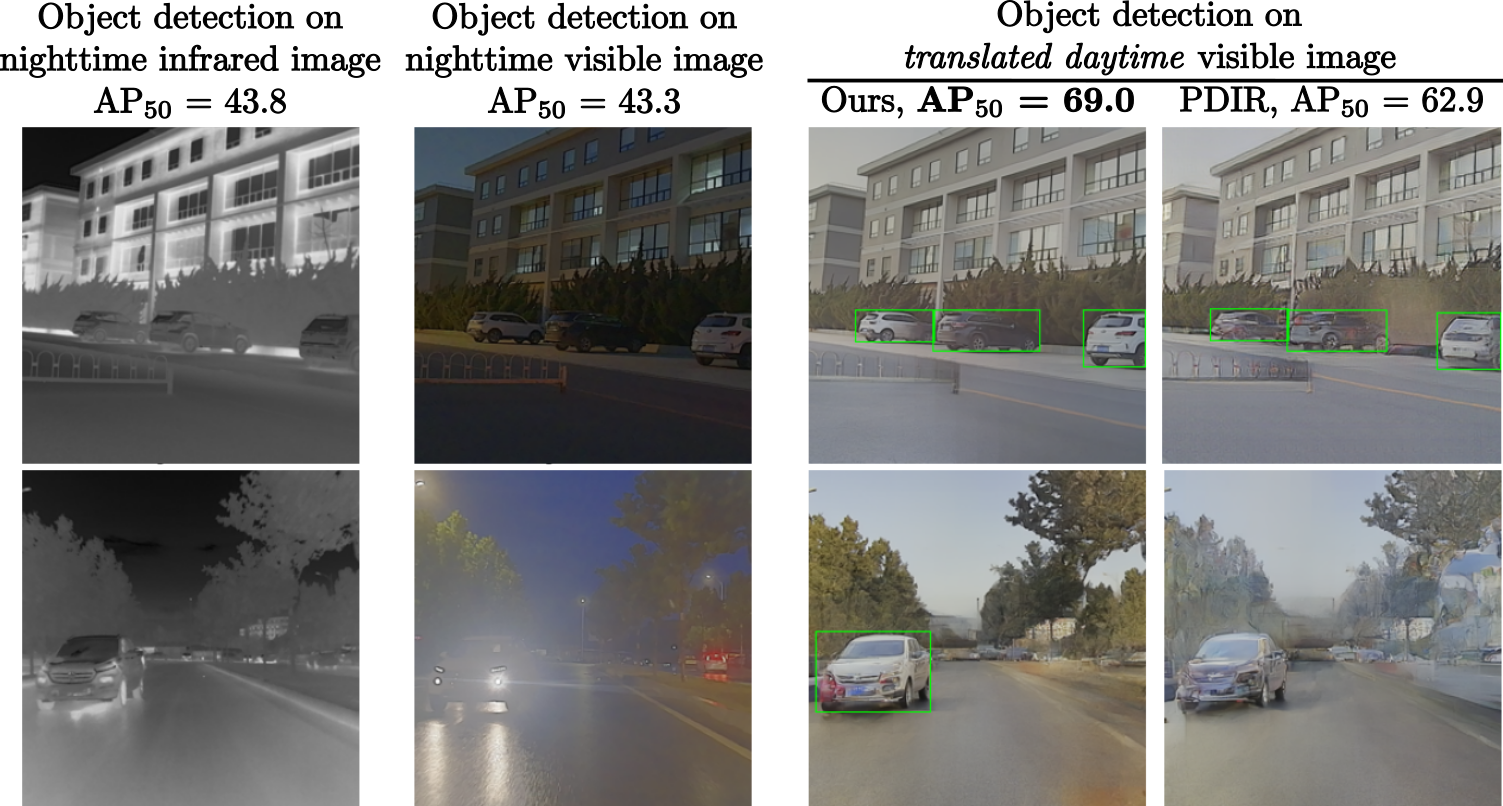}
    \caption{Day~$\rightarrow$~night object detection: We address the challenge of adapting object detectors, pretrained on daytime visible images, to nighttime scenarios. Unlike the state-of-the-art patch diffusion method PDIR~\cite{pdir},
    our method effectively preserves object structure, yielding a significant performance improvement in nighttime object detection.} 
    \label{fig:highlight}
\end{figure}

Visible light cameras, capturing rich visual information (e.g., texture, color, and structure), are widely used in various applications. Most state-of-the-art computer vision algorithms rely on RGB images, exhibiting impressive performance under clear conditions~\cite{mask2former, videoflow, yolov7}. However, in extreme environmental conditions (e.g., nighttime, rain, fog), visible images fail to provide high-quality visual information, severely impacting the performance of computer vision algorithms; see Fig.~\ref{fig:highlight}. In contrast, infrared sensors can detect thermal radiation, allowing them to sense objects emitting heat, even in adverse conditions. While infrared data offers unique advantages, its lack of visual details limits its intuitiveness for human interpretation and its applicability in practical tasks like autonomous driving or monitoring~\cite{roma}. Furthermore, the modality gap between infrared and visible images poses significant challenges for direct application of computer vision models, typically trained on visible images; see Fig.~\ref{fig:highlight}. Therefore, it is crucial to bridge this modality gap. A potential approach is to translate infrared images to visible counterparts, preserving semantic information for downstream tasks (e.g., object detection).

To address the challenge of infrared-to-visible (I2V) translation, traditional techniques~\cite{sheng2013video, gupta2012image, ulhaq2016face} rely on colour mapping functions, but require manual interventions, making them impractical in certain scenarios. Therefore, GAN-based methods have been proposed to tackle this issue~\cite{cyclegan,dualgan,dagan}, achieving more promising translation performance through adversarial training and cycle consistency strategies. However, the significant domain gap between infrared and visible modalities still remains a challenge, compromising the preservation of detailed semantic information. Furthermore, a majority of practical applications involve video data, while these image-level methods lacks mechanisms to ensure smooth inter-frame transitions. To address this, I2V-GAN methods~\cite{i2vgan, roma, cptrans} proposed temporal consistency losses in their training protocol, showing a state-of-the-art performance in I2V video translation.

Recently, diffusion models~\cite{ho2020denoising,song2019generative,song2020score} have emerged as the state of the art in generative modelling. Among these, Denoising Diffusion Probabilistic Models (DDPMs)~\cite{ddpm1, ho2020denoising} are a popular class of difussion models, 
which have been shown to outperform GANs in low-level vision tasks, such as inpainting~\cite{ldm}, super-resolution~\cite{gao2023implicit}, and restoration~\cite{pdir}. Recently, T2V-DDPM~\cite{t2vddpm} has shown promising performance in I2V translation, demonstrating the considerable promise of diffusion models for this task. 

\begin{figure}
    \centering
    \includegraphics[width=0.9\linewidth]{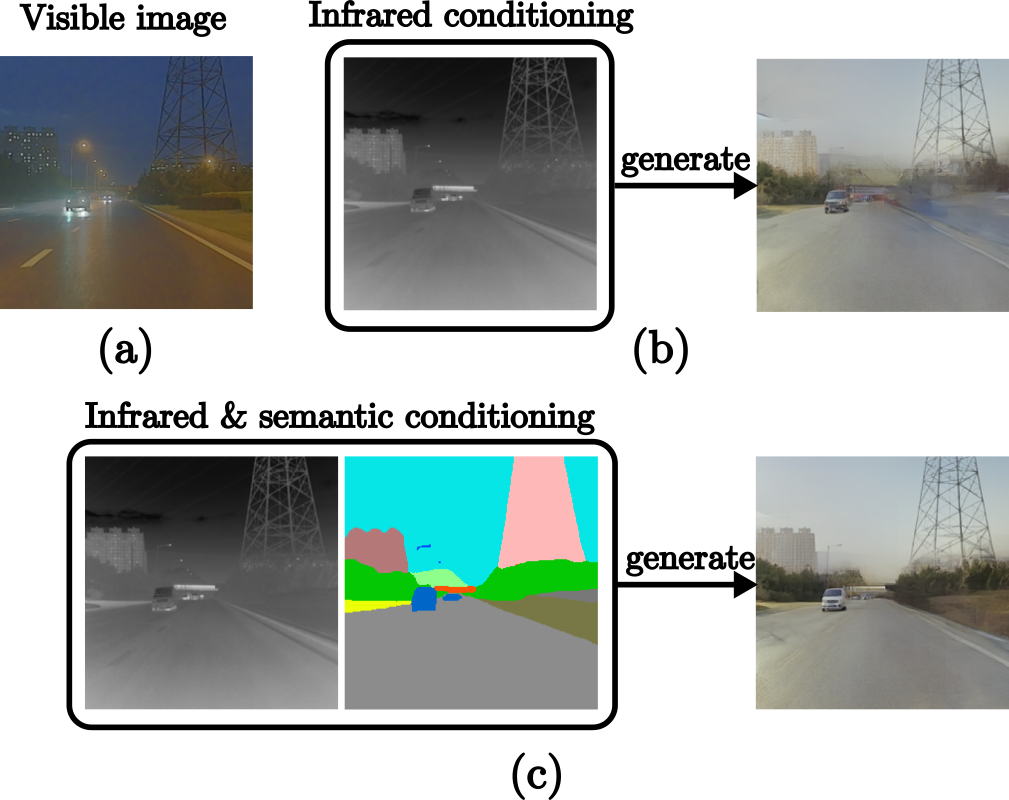}
    \caption{Benefits of semantic conditioning. \textbf{(a)} Actual nighttime visible image. \textbf{(b)} 
    Existing patch diffusion methods introduce \textit{semantic deformation} in generated visible images. \textbf{(c)} Our method effectively addresses this semantic deformation through a novel semantic conditioning strategy, enabling the generation of a visible image that faithfully  preserve  the scene's structure.}
    \label{fig:sem_benefit}
\end{figure}

However, DDPMs are designed for full-size images, making them computationally intensive in training. To alleviate this, patch diffusion models~\cite{pdir, multidiffusion, patchdiffusion} have been introduced, offering a promising solution. Despite showing great potential in image synthesis tasks, two research gaps remain in the existing patch diffusion frameworks. \textbf{(i) Lack of emphasis on semantic preservation:} Although existing patch diffusion methods demonstrated impressive perceptual quality, our finding reveals that it tends to deform semantic structures of small objects in complex scenes, as shown in  Figs.~\ref{fig:highlight} and~\ref{fig:sem_benefit}.  
Crucially, preserving semantic structure is vital for downstream tasks like object detection, highlighting the importance of integrating additional semantic conditioning to patch diffusion frameworks. \textbf{(ii) Temporal inconsistency:} Direct application of patch diffusion to video translation yields unsmooth inter-frame transitions; see Fig.~\ref{fig:quantitative_smoothness}, resulting in generated videos that are unrealistic. Therefore, guaranteeing temporal consistency is essential in the generation of consecutive visible frames. To this end,
this paper makes following contributions:
\begin{enumerate}[label=(\roman*)]
    \item We introduce a novel semantic-conditioning patch diffusion model, 
    using a foundational segmentation model to preserve structural information in generated visible images. Our method yields dual benefits: enhanced visual fidelity and significantly improved object detection performance; see Figs.~\ref{fig:highlight} and~\ref{fig:sem_benefit}.
    \item We propose a novel temporal blending module that uses dense correspondences, established by optical flow, to guide the direction of denoising trajectories. This maintains temporal consistency, enabling smooth inter-frame transitions; see Fig.~\ref{fig:quantitative_smoothness}.
    \item Experiment shows that our method outperforms state of the art by 35.3\% in FVD for I2V video translation and by 6.1\% in AP$_{50}$ for day~$\rightarrow$~night object detection.
    
\end{enumerate}

\section{Related work}
\subsection{Infrared-to-visible translation} 
I2V translation has emerged as a key research area due to its great potential in enhancing nighttime vision. One popular strategy is infrared-visible fusion~\cite{m3fd,ddfm,zhao2024equivariant} which combines the advantages of both modalities.
However, the lack of visual details of visible images in low-light conditions tends to compromise the effectiveness of fused images. Another approach is colourization, which involve learning a colour mapping function to generate visible images~\cite{suarez2018near,suarez2017infrared,suarez2018learning}. Nevertheless, these methods are prone to details distortion and, by neglecting temporal consistency, inherently limit the I2V video translation performance.

Recently, generative model have shown a state-of-the-art performance in I2V translation. For instance, Pix2Pix~\cite{pix2pix} employed deep neural networks in the GAN framework~\cite{gan}, while CycleGAN~\cite{cyclegan} proposed cycle consistency to ensure consistent forward and reverse mappings. Also, geometric-constrained GAN methods~\cite{fu2019geometry,johnson2016perceptual,mechrez2018contextual} further enhanced the structural coherence by incorporating geometric consistency. However, these methods primarily focused on preserving spatial information, leading to temporal inconsistencies when directly applied to video data. To address this, GAN-based video translation methods~\cite{recyclegan, mocyclegan, unsuprecycle} enforced smooth transitions between frames using motion-guided mechanisms. Nevertheless, the significant domain gap between infrared and visible modalities limited their capability to produce realistic visible videos. Therefore, I2V-GAN~\cite{i2vgan} and its variants~\cite{roma,cptrans} have been proposed, showing promising results in the task.

Meanwhile, diffusion models have demonstrated its impressive performance in generative modelling, outperforming GAN-based methods in image synthesis tasks. Recently, T2V-DDPM~\cite{t2vddpm} was introduced to address I2V translation, but its neglect of temporal constraints resulted in poor performance in video data. In contrast, our \ourmethod preserves both temporal and semantic consistency between infrared and visible modalities, yielding a significant improvement in generating realistic visible videos. 

\subsection{Patch diffusion models}
Patch diffusion models have been shown to be more computationally efficient than full-resolution diffusion models~\cite{hpdm, patchdiffusion, multidiffusion}.
Previous works~\cite{multidiffusion,kim2023collaborative,zheng2024any} primarily focused on leveraging foundational text-to-image generators (e.g., Stable Diffusion~\cite{stablediffusion}) and improving sampling strategies, allowing them to generate images at higher resolution than their original training data. Recent works explored the patch-level training for diffusion models. PDIR~\cite{pdir} trained a U-Net on randomly cropped image patches and blended patch-level noise predictions to reconstruct full-resolution noise during inference. PatchDiffusion~\cite{patchdiffusion} proposed to progressively increase the patch size over the training iterations, whereas HDPM~\cite{hpdm} introduced a coarse-to-fine training strategy, ensuring spatial consistency by extracting fine-level patches within a coarse-level patch. 

Unlike existing patch diffusion methods primarily focusing on visual realism, our approach prioritises a dual objective: visual quality and downstream task performance (e.g., object detection). To this end, we introduce semantic conditioning to preserve object structure and temporal blending to ensure smooth frame transitions.

\section{Background}
\subsection{Denoising diffusion probabilistic models}
DDPMs~\cite{ho2020denoising,ddim} are a class of generative models which learn data distribution using a Markovian process. 

During training, DDPMs define a  \textit{diffusion process}, which corrupts an image $X_0$ to a white Gaussian noise $X_T \sim \cN(0, \bI)$ in $T$ time steps. The operation for each step is defined as: 
$q\left(X_t \, | \, X_{t-1}\right) = \cN \left(X_t \, ; \, \sqrt{1-\beta_t}X_{t-1}, \beta_t \bI \right)
    \label{eq:ddpm:forward}$,
where, $\left\{\beta_t \right\}_{t=0}^T$ is a predefined variance schedule. The state $X_t$ can also be sampled from the clean image $X_0$: $q \left(X_t \, | \, X_0 \right) = \cN \left(X_t \, ; \, \sqrt{\bar{\alpha}_t} X_0, \left(1 - \bar{\alpha_t}\right) \bI \right)$, which can be represented in the closed-form
\begin{equation}
    X_t = \bar{\alpha_t}X_0  + \left(1 - \bar{\alpha_t}\right)\epsilon
    \label{eq:sample_Xt_from_X0}
\end{equation}
where, $\epsilon \sim \cN\left(0, \bI \right)$, $\bar{\alpha_t} = \prod_{s=1}^t \alpha_i$, and $\alpha_t = 1-\beta_t$. 


During sampling, DDPMs define a \textit{denoising process}. Starting from a random noise $X_T \sim \cN\left(0, \bI\right)$, a clean image $X_0$ can be sampled through a series of denoising steps
\begin{equation}
    X_{t-1} = \frac{1}{\sqrt{\alpha_t}} \left(X_t - \frac{\beta_t}{\sqrt{1-\bar{\alpha_t}}} \bepsilon_\theta\left(X_t, t\right) \right) + \sigma_t \bz
\end{equation}
where, $\bz \sim \cN(0, \bI)$ makes denoising steps stochastic. The noise predictor $\bepsilon_\theta$ parameterised by a neural network with weight $\theta$ can be trained using a simplified training objective

\begin{equation}
    \bbE_{X_0, t, \epsilon \sim \cN\left(0, \bI\right)} \left[ \|\epsilon - \bepsilon_\theta\left(X_t, t \right) \|^2\right],
\end{equation}
where $X_t$ can be computed from the clean image $X_0$ and $\epsilon$ using Eq.~\eqref{eq:sample_Xt_from_X0}. 

\subsection{Conditional denoising diffusion models}
Given a conditional image $Y$, we are interested in sampling a new image $X_0$ preserving certain attributes of $Y$. To this end, the neural network is conditioned by the conditional image at all timesteps $t$, where the training objective is expressed by~\cite{saharia2022image}

\begin{equation}
    \bbE_{X_0, t, \epsilon \sim \cN\left(0, \bI\right)} \left[ \|\epsilon - \bepsilon_\theta\left(X_t, Y, t \right) \|^2\right]
\end{equation}
Once the neural network $\bepsilon_\theta$ is trained, given an arbitrary conditional image $Y^\prime$, the new image $X^\prime_0$ can be sampled through a sequence of denoising steps

\begin{equation}
    X^\prime_{t-1} = \frac{1}{\sqrt{\alpha_t}} \left(X^\prime_t - \frac{\beta_t}{\sqrt{1-\bar{\alpha_t}}} \bepsilon_\theta\left(X^\prime_t, Y^\prime, t\right) \right) + \sigma_t \bz
\end{equation}
where, the variance schedule is unchanged from the training.

\begin{figure*}
    \centering
    \includegraphics[width=1.0\linewidth]{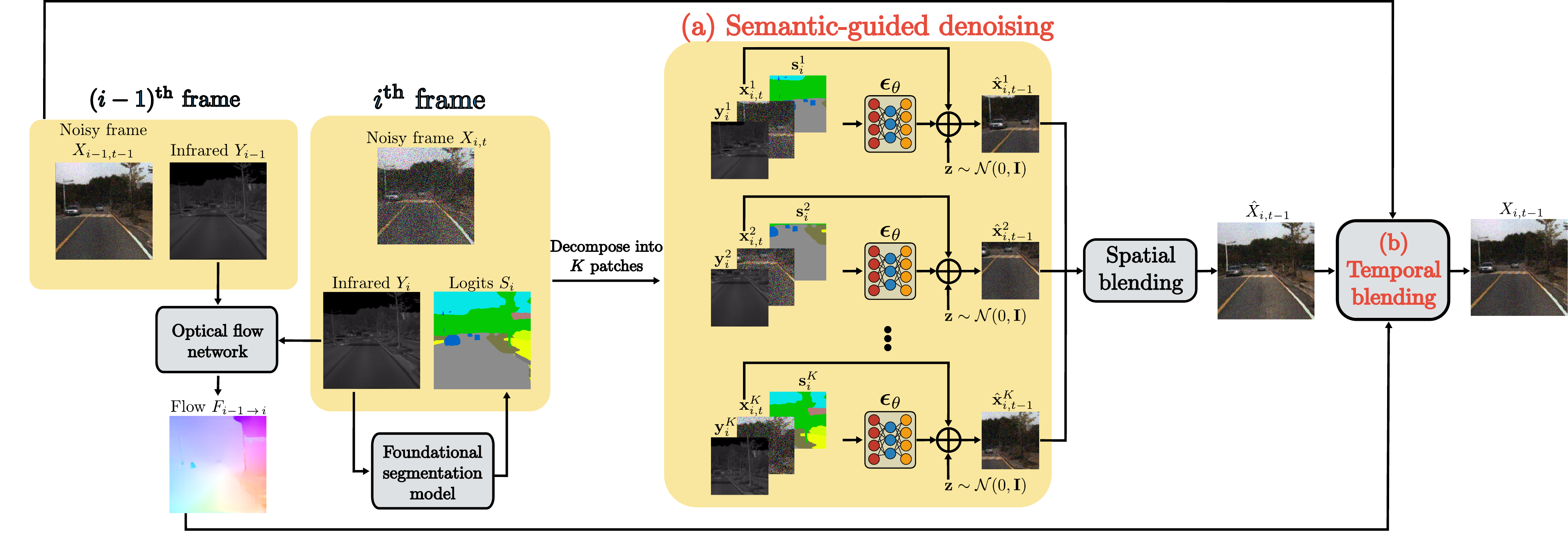}
   \caption{Overview of \ourmethod at timestep $t$: Our contribution consist of two components. \textbf{(a) Semantic-guided denoising:} By leveraging the foundational model, we estimate the segmentation logits $S_i$ for the infrared image $Y_i$, which injects additional semantic knowledge into the denoising process. This additional semantic condition ensures the generated visible image faithfully reproduces the scene's structural information.
   \textbf{(b) Temporal blending:} We leverage a pretrained optical flow network to estimate the flow $F_{i-1 \, \rightarrow \, i}$ from consecutive infrared images, which serves as a guidance for the denoising trajectory's direction. This ensures that the generated frame remains temporally consistent with the preceding frame.}
    \label{fig:i2v-pdm}
\end{figure*}

\section{Methodology}
\label{sec:method}

\paragraph{Problem statement.} 
Let a training dataset $\cD = \{X_j, Y_j\}_{j=1}^M$ of $M$ pairs, each containing a visible image $X_j \in \bbR^{H\times W \times 3}$ and a corresponding infrared image $Y_j \in \bbR^{H\times W}$. The objective is to train a translation model, which can translate an infrared video $\cY = \left\{Y_i\right\}_{i=1}^N$ of $N$ frames to a visible video $\cX = \left\{X_i\right\}_{i=1}^N$. Note that $\cX$ must accurately preserve the scene's structure of $\cY$. 

\paragraph{Overview} Our method, which is illustrated in Fig.~\ref{fig:i2v-pdm}, contains two key components: \textit{(i) Semantic-guided denoising} to accurately preserve the scene's structure and \textit{(ii) temporal blending} to ensure the temporal consistency between consecutive frames; see Fig.~\ref{fig:benefits_sem_tb} for the benefits of each component. The training of \ourmethod will be provided in Sec.~\ref{sec:train} and the I2V video generation will be elaborated in Sec.~\ref{sec:gen}.


\subsection{Training with semantic conditioning}
\label{sec:train}
Existing patch diffusion models~\cite{pdir,t2vddpm,hpdm} are prone to distort object structures, as shown in Fig.~\ref{fig:sem_benefit}. This phenomenon, which we term \textit{semantic deformation}, happens because data-hungry diffusion models typically require a vast amount of training data to capture structural details of small objects within complex scenes. For instance, GAIA-1~\cite{gaia} requires training on 420M images to achieve desired performance in complex scenarios. 

Therefore, to preserve semantic structures, we propose to leverage pre-trained knowledge of a foundational segmentation model, and extract semantic segmentation $S_j \in \bbR^{H\times W \times L}$ of the input infrared image $Y_j$. Here, $L$ represents the number of object categories, pre-learned by the segmentation model.
The segmentation logits $S_j$, output by the softmax, provides a probability distribution over object categories. This distribution indicates the likelihood of a pixel belonging to each specific category. 

Unlike prior works using segmentation masks~\cite{smartbrush,chen2024towards,park2024shape}, we propose to use segmentation logits. This is driven by the hypothesis that foundational models are typically trained on visible data. Hence, directly applying it to infrared data could introduce prediction errors due to the domain gap. To alleviate this, we use segmentation logits as soft predictions, reducing the impact of potential errors associated with hard predictions like segmentation masks. We showed that using segmentation logits significantly outperforms using masks; see Fig.~\ref{fig:masks_vs_logits}.


Training $\bepsilon_\theta$ for \ourmethod is similar to regular DDPMs. However, instead of using full-size images, we use patches randomly subsampled on a pair 
in $\cD$. Specifically, let $p \times p$ be the size of the patch, the top-left coordinates $(u,v)$ are randomly generated. Using $(u,v)$, the patch tuple $\{\bx_j, \by_j, \bs_j \}$ is cropped from $\{X_j, Y_j, S_j \}$, where $\bx_j \in \bbR^{p \times p \times 3}$, $\by_j \in \bbR^{p \times p}$, and $\bs_j \in \bbR^{p \times p \times L}$. The model $\bepsilon_\theta$ is then trained using the following objective
\begin{equation}
    \bbE_{\bx_j, \by_j, \bs_j, t, \epsilon \sim \cN\left(0, \bI\right)} \left[ \|\epsilon - \bepsilon_\theta\left(\bx_{j,t}, \by_j, \bs_j, t \right) \|^2\right],
\end{equation}
where, $\bx_{j,t} = \bar{\alpha_t}\bx_j  + \left(1 - \bar{\alpha_t}\right)\epsilon$. 

\subsection{Temporally-consistent video generation}
\label{sec:gen}
For each $i^\text{th}$ infrared frame $Y_i$ of the video $\cY$, we use the foundational segmentation model to extract the segmentation logits $S_i$, which will serve as the semantic conditioning in generating the corresponding visible frame $X_i$. 

Starting with a random noise $X_{i,T} \sim \cN\left(0, \bI\right)$, the visible image $X_i$ is generated through a series of denoising steps. Each denoising step, as illustrated in Fig.~\ref{fig:i2v-pdm}, contains two essential components: semantic-guided denoising and temporal blending.

\paragraph{Semantic-guided denoising} Unlike prior patch diffusion methods~\cite{pdir, patchdiffusion, multidiffusion, hpdm}, which overlook semantic information during denoising, we explicitly integrate semantic conditioning into the denoising process. 

At each timestep $t$, we firstly decompose the tuple $\{X_{i,t}, Y_i, S_i\}$ into $K$  patches, denoted as $\{\bx^k_{i,t}, \by^k_i, \bs^k_i\}_{k=1}^K$. Our patch decomposition is inspired by PDIR~\cite{pdir}. In particular, we partition an image into a grid of cells, each with size of $r \times r$. We then crop a set of patches by sliding a $p \times p$ window horizontally and vertically across the grid, with a step of $r$.


Next, using the semantic logits $\bs^k_i$ and the infrared patch $y^k_i$, each noisy patch $\bx_{i,t}^k$ is denoised as follows

\begin{equation}
    \hat{\bx}^k_{i, t-1} = \frac{1}{\sqrt{\alpha_t}} \left(\bx^k_{i,t} - \frac{\beta_t}{\sqrt{1-\bar{\alpha_t}}} \bepsilon_\theta\left(\bx^k_{i,t}, \by^k_i, \bs^k_i, t\right) \right) + \sigma_t \bz
    \label{eq:simultaneous_denoising}
\end{equation}
where, $\bz \sim \cN(0, \bI)$. Our denoising process is guided by two key inputs to the model $\bepsilon_\theta$: the infrared image $\by^k_i$ and the semantic logits $\bs^k_i$. By incorporating semantic logits $\bs^k_i$, we ensure that the generated visible image will preserve the scene's structural integrity, as described by the semantic information in $\bs^k_i$; see Fig.~\ref{fig:sem_benefit}.

Finally, given a set $K$ of denoised patches $\{\hat{\bx}^k_{i,t-1} \}_{k=1}^K$, we perform the spatial blending to reconstruct a full-size image $\hat{X}_{i,t-1} \in \bbR^{W \times H \times 3}$. Our spatial blending draws inspiration from PDIR~\cite{pdir}. Specifically, each pixel of $\hat{X}_{i,t-1}$ is computed as a weighted average, where the weight assigned to each pixel is determined by the number of overlapping patches at that pixel.

\begin{algorithm}[h]\centering
	\begin{algorithmic}[1]
	\REQUIRE The $i^\text{th}$ frame $\hat{X}_{i,t-1}$, the $(i-1)^\text{th}$ frame $X_{i-1,t-1}$, flow $F_{i-1 \, \rightarrow \, i}$, weight $w_t$, and decay factor $\omega$

        \STATE Initialise correspondences $\cM_{i-1,i} = \emptyset$ \label{algo:temporal_blend:init_corr}
        \FOR{every coordinate $(u,v)$ in the $(i-1)^\text{th}$ frame}
            \STATE $(u^\prime, v^\prime)  = \texttt{round}\big((u,v) + F_{i-1 \, \rightarrow \, i}[u,v]\big)$ \label{algo:temporal_blend:coordinate_displacement}
            \STATE $\cM_{i-1,i} \leftarrow \cM_{i-1,i} \, \cup \, (u, v, u^\prime, v^\prime)$
        \ENDFOR
                
        \STATE $\cM_{i-1,i} \leftarrow$ \texttt{geometric\_verification}($\cM_{i-1, i}$). \label{algo:temporal_blend:geometric_verify}
        \STATE Decay weight $w_{t-1} \leftarrow w_t.\omega$ \label{algo:temporal_blend:decay}
        \STATE Initialise $X_{i, t-1}$ $\leftarrow \hat{X}_{i, t-1}$ 
        \FOR{every correspondence $(u, v, u^\prime, v^\prime)$ in $\cM_{i-1,i}$} \label{algo:temporal_blend:linear_comb_start}
            \STATE $X_{i, t-1}[u^\prime, v^\prime] \leftarrow (1-w_{t-1}).\hat{X}_{i, t-1}[u^\prime, v^\prime] + w_{t-1}. X_{i-1, t-1}[u,v]$  \label{algo:temporal_blend:linear_comb}
        \ENDFOR \label{algo:temporal_blend:linear_comb_end}
	\RETURN Blended image $X_{i, t-1}$ and weight $w_{t-1}$
	\end{algorithmic}
	\caption{Temporal blending.}
	\label{algo:temporal_blend}
\end{algorithm}

\paragraph{Temporal blending} Algo.~\ref{algo:temporal_blend} illustrates our temporal blending. Given the flow $F_{i-1 \, \rightarrow \, i}$, the correspondences $\cM_{i-1,i}$ between $(i-1)^\text{th}$ and $i^\text{th}$ frames are estimated (line~\ref{algo:temporal_blend:init_corr}-\ref{algo:temporal_blend:geometric_verify}). Specifically, for every coordinate in the $(i-1)^\text{th}$ frame, we estimate the its correspondence in the $i^\text{th}$ frame using the optical flow velocities (line~\ref{algo:temporal_blend:coordinate_displacement}). To further improve the correspondences quality, we apply geometric verification through the fundamental matrix estimation inside RANSAC iterations (line~\ref{algo:temporal_blend:geometric_verify}). Finally, at every correspondence, the pixel value of blended image $X_{i,t-1}$ is a linear combination between $\hat{X}_{i,t-1}$ and $X_{i-1, t-1}$ (line~\ref{algo:temporal_blend:linear_comb}). Note that the weight $w$ in this linear combination is decayed at every denoising timestep (line~\ref{algo:temporal_blend:decay}), see Fig.~\ref{fig:decay_effects} for the benefits of decay. 

\begin{figure}[h]
    \centering
    \includegraphics[width=1.0\linewidth]{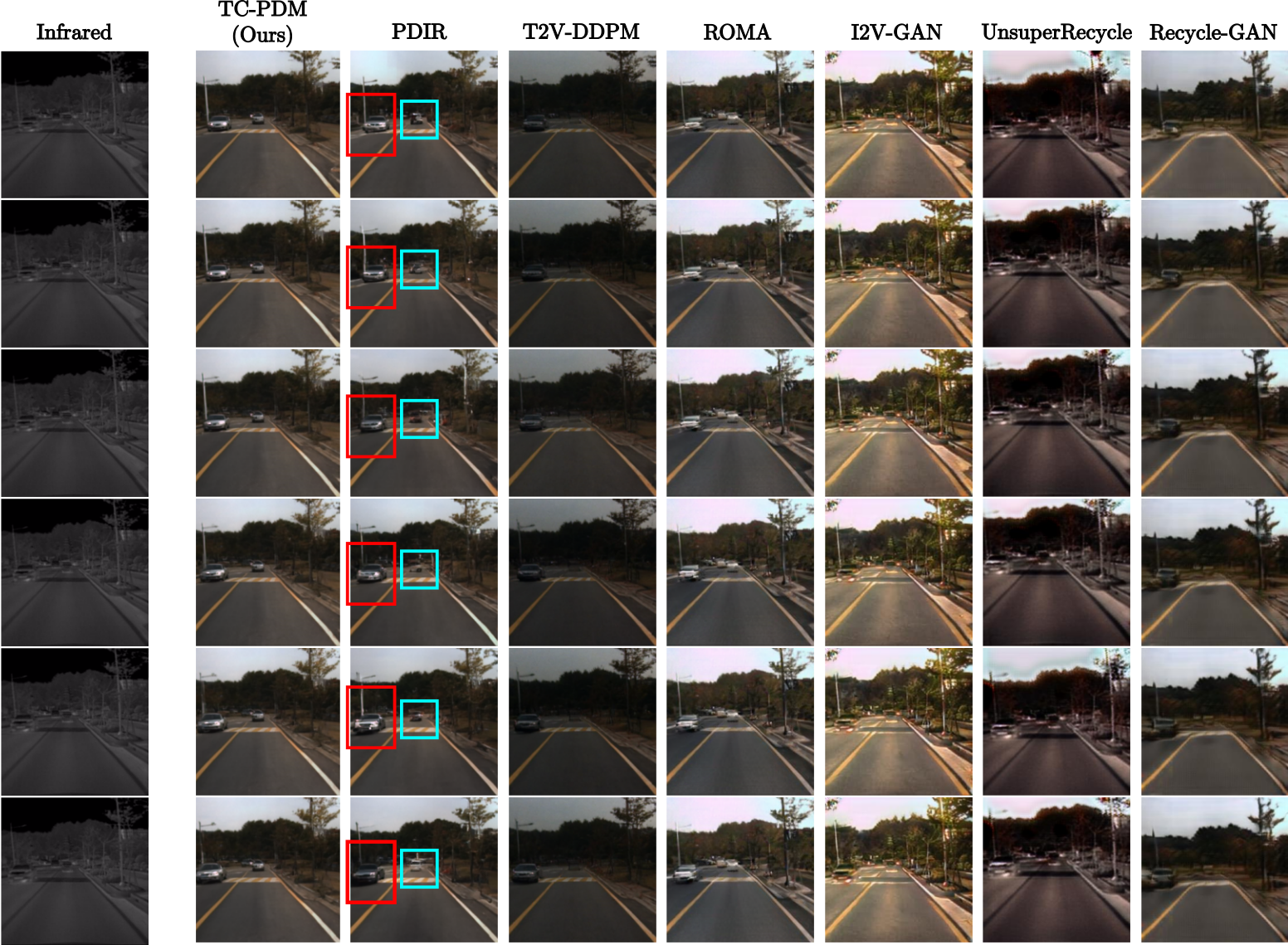}
    \caption{The neglect of temporal consistency in patch diffusion method PDIR leads to structurally-inconsistent objects across frames (highlighted). By contrast, our method uses a novel temporal blending module to maintain consistent object structures across frames, achieving temporal continuity.}
    \label{fig:quantitative_smoothness}
    
\end{figure}

\subsection{Implementation} 
Our foundational segmentation model consists of a DINOv2 backbone~\cite{dino} with a Mask2Former head~\cite{mask2former}. The Mask2Former was trained on the ADE20K dataset~\cite{ade20k} comprising $L = 150$ object categories. In addition, we employed the VideoFlow model~\cite{videoflow}, trained on the Sintel dataset~\cite{sintel}, as our optical flow network.

\begin{table*}
    \centering
    \footnotesize
    \begin{tabular}{lcccccccccccc}
        \toprule
        & \multicolumn{4}{c}{\textbf{V000}} & \multicolumn{4}{c}{\textbf{V001}} & \multicolumn{4}{c}{\textbf{V002}} \\
         \cmidrule(lr){2-5}  \cmidrule(lr){6-9}  \cmidrule(lr){10-13} 
          & \textbf{PSNR}$\uparrow$  & \textbf{SSIM}$\uparrow$ &  \textbf{FID}$\downarrow$ & \textbf{FVD}$\downarrow$ & \textbf{PSNR}$\uparrow$  & \textbf{SSIM}$\uparrow$ &  \textbf{FID}$\downarrow$ & \textbf{FVD}$\downarrow$ & \textbf{PSNR}$\uparrow$  & \textbf{SSIM}$\uparrow$ &  \textbf{FID}$\downarrow$ & \textbf{FVD}$\downarrow$ \\
          \midrule
          Recycle-GAN & 15.6 & 0.51 & 146.5 & 332.3 & 12.5 & 0.43 & 178.6 & 424.8 & 13.9 & 0.48 &  143.2 & 400.8 \\
         UnsuperRecycle & 13.1 & 0.45 & 185.4 & 383.6 & 12.2 & 0.42 & 187.8 & 538.3 & 13.4 & 0.46 & 170.1 & 
         354.6 \\
         I2V-GAN & 15.8 & 0.54 & 121.1 & 218.9 & 13.6 & 0.45 & 148.0 & 278.8 & 14.3 & 0.46 & 127.7 & 267.9 \\
         ROMA & 15.8 & 0.55 & 80.1 & 112.9 & 14.4 & 0.50 & 83.8 & 203.1 & 15.8 &  0.55 & 72.5 & 171.5 \\
         T2V-DDPM & 12.6 & 0.57 & 66.6 & 273.1 & 13.5 & 0.54 & 65.4 & 287.3 & 14.2 & 0.58 & 56.1 & 235.2\\
         PDIR & 18.1 & 0.65 & \textbf{48.7} & 617.0 & \textbf{14.9} & 0.56 & \textbf{55.2} & 727.6 & 16.3 & 0.61 & \textbf{50.5} & 568.4\\
         \textbf{\ourmethod (Ours)} & \textbf{18.2} & \textbf{0.66} & 51.8 & \textbf{82.1} & 14.8 & \textbf{0.57} & 60.4 & \textbf{125.2} & \textbf{16.4} & \textbf{0.62} & 52.0 & \textbf{88.0}\\
         \bottomrule
    \end{tabular}
    \caption{Comparison of different translation methods on the KAIST dataset for infrared-to-visible video translation.}
    \label{tab:vid_trans}
\end{table*}

The network $\bepsilon_\theta$ was implemented using U-Net~\cite{unet} based on WideResNet architecture~\cite{wideresnet}, with self-attention at $16 \times 16$ resolution~\cite{attention} and group normalisation~\cite{groupnorm}. Sinusoidal positional embedding~\cite{attention} was used for the embedding of timestep $t$, which was then provided as input to each residual block of U-Net. For infrared and semantic conditioning, the infrared patch $\by^k_i$ and semantic patch $\bs^k_i$ are concatenated with the noisy visible patch $\bx^k_{i,t}$ on a channel-wise basis, resulting in $156$-D input channels. Also, we set patch size $p=64$, cell size $r=16$, weight $w_T = 1$, and decay factor $\omega = 0.9$.

During training, we randomly sampled $8$ images and cropped $4$ patches from each image, yielding the training batch size of $32$. The model $\bepsilon_\theta$ was trained in $2$M iterations (KAIST) or $1.2$M iterations (M3FD) using four V100 GPUs and Adam optimiser with learning rate of $0.00002$. For a stable training, we employed an exponential moving average with a momentum of $0.999$~\cite{song2020improved}.

\section{Experiments}
\ourmethod was evaluated in 2 tasks: I2V video translation (Sec.~\ref{sec:exp:i2v}) and day~$\rightarrow$~night object detection (Sec.~\ref{sec:exp_obj_det}).



\begin{figure*}[h]
    \centering
    \includegraphics[width=1.0\linewidth]{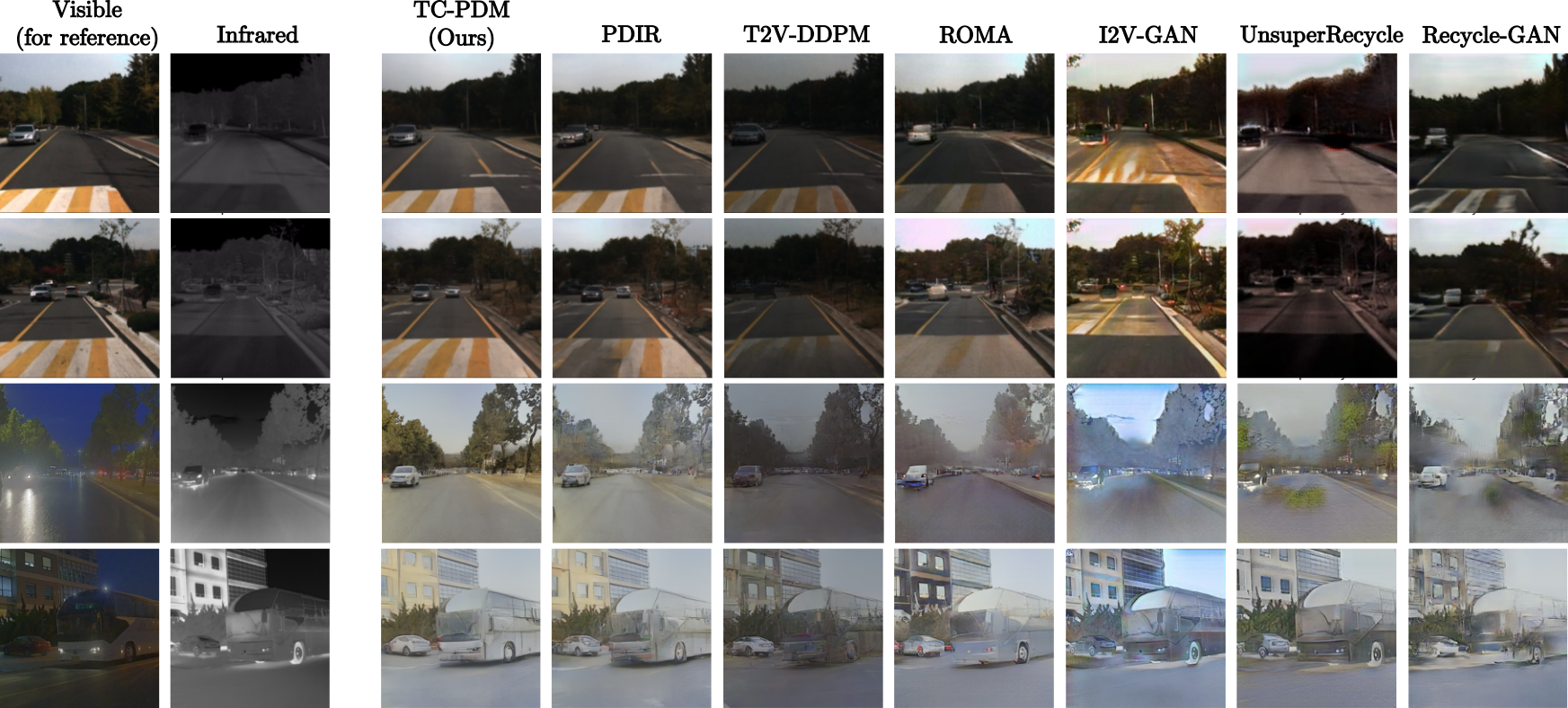}
    \caption{Qualitative comparison of different translation methods on the KAIST (rows 1-2) and M3FD (rows 3-4) datasets. In KAIST, our \ourmethod generates visible images that are perceptually and structurally more similar to the \textit{actual} visible images. Similarly, on M3FD, our generated ``fake" visible images exhibit more details and preserve the structure of dynamic objects like cars. More qualitative results can be found in Sec.~\ref{supp:sec:i2v_more_results} in supplementary material.}
    \label{fig:quantitative_quality}
\end{figure*}

\subsection{Infrared-to-visible translation}
\label{sec:exp:i2v}

\paragraph{Dataset} we used KAIST dataset~\cite{kaist}, which provides about 95k visible-infrared image pairs. Due to the limitation of our computational resources, we selected all videos of set00 for training and videos V000-002 of set06 for testing, resized and center-cropped all images to 256$\times$256. This results in 17,493 infrared-visible training pairs and 5,888 infrared-visible testing pairs. 

\paragraph{Baselines} We compared \ourmethod to four GAN methods: Recycle-GAN~\cite{recyclegan}, I2V-GAN~\cite{i2vgan} , UnsuperRecycle~\cite{unsuprecycle}, and ROMA~\cite{roma}; as well as two diffusion methods: PDIR~\cite{pdir} and T2V-DDPM~\cite{t2vddpm}.

\begin{figure*}
    \centering
    \subfloat[]
    {
        \begin{tikzpicture}
            \begin{axis}
                [width=5.8cm,
                height=5.5cm,
                grid=major,
                ylabel = FVD,
                ylabel near ticks,
                ymin=0, ymax=1200,
                ytick={100,300,500,700, 900, 1100},
                yticklabels={100,300,500,700, 900, 1100},
                xlabel = Training iterations,
                xlabel near ticks,
                xtick={300, 1000, 1500, 2000},
                xticklabels={300K, 1M, 1.5M, 2M},
                legend style={at={(0.20,0.75)},anchor=west,font=\scriptsize}, 
                legend cell align={left}]
                
                \addplot[mark=triangle*,black, line width=1.5pt] table[x=train_iters,y=fvd, col sep=comma] {data/kaist_our_method_wo_logits_time.csv};
                \addlegendentry{\ourmethod w/o SC \& TB}
        
                \addplot[mark=*,green, line width=1.5pt] table[x=train_iters,y=fvd, col sep=comma] {data/kaist_our_method_wo_logits.csv};
                \addlegendentry{\ourmethod w/o SC}
        
                \addplot[mark=pentagon*,blue, line width=1.5pt] table[x=train_iters,y=fvd, col sep=comma] {data/kaist_our_method_wo_time.csv};
                \addlegendentry{\ourmethod w/o TB}
        
                \addplot[ mark=square*,red, line width=1.5pt] table[x=train_iters,y=fvd, col sep=comma] {data/kaist_our_method.csv};
                \addlegendentry{\ourmethod}
            \end{axis}
        \end{tikzpicture}
        \label{fig:benefits_sem_tb}
    }
    \subfloat[][]
    {
        \begin{tikzpicture}
            \begin{axis}[
                width=5.8cm,
                height=5.5cm,
                grid=major,
                ylabel = FVD,
                ylabel near ticks,
                ymin=0, ymax=600,
                ytick={100,200,300,400, 500, 600},
                yticklabels={100,200,300,400, 500, 600},
                xlabel = Decay factor $\omega$,
                xlabel near ticks,
                xtick={0.0, 0.2, 0.4, 0.6, 0.8, 1.0},
                xticklabels={0.0, 0.2, 0.4, 0.6, 0.8, 1.0}]
                
                \addplot[mark=square,red, line width=1.5pt]
                table[x=decay,y=fvd, col sep=comma] {data/decay.csv};
            \end{axis}
        \end{tikzpicture}
        \label{fig:decay_effects}
    }
    \subfloat[]
    {
        \begin{tikzpicture}
            \begin{axis}
                [width=5.8cm,
                height=5.5cm,
                grid=major,
                ylabel = FVD,
                ylabel near ticks,
                ymin=0, ymax=400,
                ytick={100,200, 300, 400},
                yticklabels={100,200, 300, 400},
                xlabel = Training iterations,
                xlabel near ticks,
                xtick={300, 1000, 1500, 2000},
                xticklabels={300K, 1M, 1.5M, 2M},
                legend style={at={(0.09,0.79)},anchor=west,font=\scriptsize}, 
                legend cell align={left}]
    
                 \addplot[mark=*,green, line width=1.5pt] table[x=train_iters,y=fvd, col sep=comma] {data/kaist_our_method_wo_logits.csv};
                \addlegendentry{\ourmethod w/o SC}
                
                \addplot[mark=square*,blue, line width=1.5pt] table[x=train_iters,y=fvd, col sep=comma] {data/kaist_our_method_with_masks.csv};
                \addlegendentry{\ourmethod with SC on masks}
    
                 \addplot[mark=pentagon*,red, line width=1.5pt] table[x=train_iters,y=fvd, col sep=comma] {data/kaist_our_method.csv};
                \addlegendentry{\ourmethod with SC on logits}
    
            \end{axis}
        \end{tikzpicture}
         \label{fig:masks_vs_logits}
    }
    
    \caption{Ablation studies using video V002: \textbf{(a)} The benefits of semantic conditioning (SC) and temporal blending (TB) in the translation performance. \textbf{(b)} The effects of various decay factors $\omega$ to the translation quality. \textbf{(c)} Comparison of semantic conditioning (SC) on segmentation masks vs. logits.}
\end{figure*}
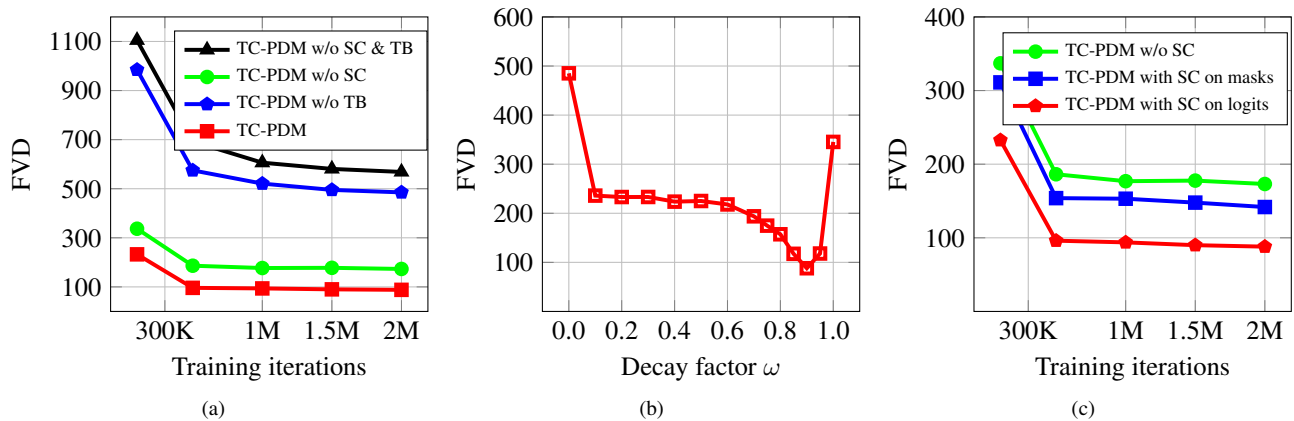

\paragraph{Metrics} 
we reported four metrics. 
\begin{itemize}
    \item Peak Signal-to-Noise Ratio (\textbf{PSNR})~\cite{psnr} evaluates the translation quality---higher PSNR indicating less distortion in translated images.
    \item Structural Similarity (\textbf{SSIM})~\cite{ssim} measures structural information in translated images, with larger SSIM indicating higher similarity to actual images.
\end{itemize}

Following~\cite{pdir, valanarasu2022transweather}, PSNR and SSIM were computed on the luminance channel Y of the YCbCr color space.
\begin{itemize}
    \item Fr\'echet Inception distance (\textbf{FID})~\cite{fid} estimates the domain gap between real and translated images using Inception-v3 features~\cite{inceptionmodel}, with lower FID scores indicating a smaller domain gap.
    \item Fr\'echet video distance (\textbf{FVD})~\cite{unterthiner2018towards} extends FID to measure the domain gap for video data. We used the recent FVD implementation~\cite{fvd} with video model VideoMAE-v2~\cite{wang2023videomae}.
\end{itemize}

\paragraph{Comparison to other approaches} In general, \ourmethod significantly outperforms existing methods across key metrics: PSNR, SSIM, LPIPS, and FVD; see Tab.~\ref{tab:vid_trans} and Fig.~\ref{fig:quantitative_quality}. More qualitative results can be found in Sec.~\ref{supp:sec:i2v_more_results} in the supplementary material. 

While PDIR narrowly leads in FID score, our \ourmethod achieves a significant performance in FVD, showing its efficency in capturing both translation quality and temporal consistency. This advantage is attributed to \ourmethod's incorporation of semantic conditioning and temporal blending in our framework; see Figs.~\ref{fig:quantitative_smoothness} and~\ref{fig:quantitative_quality}.

We also conducted an additional user study for comparison; see Sec.~\ref{supp:sec:i2v_more_results} in the supplementary material.

\paragraph{Benefits of semantic conditioning and temporal blending} We investigated the benefits of incorporating semantic conditioning $\bs_i^k$ (see Eq.~\eqref{eq:simultaneous_denoising}) and the temporal blending (see Algo.~\ref{algo:temporal_blend}), and found that their combination significantly improves the performance; see Fig.~\ref{fig:benefits_sem_tb}. In particular, without semantic conditioning and temporal blending, \ourmethod performs poorly in FVD across all training iterations. In contrast, incorporating either one substantially improves FVD scores, with temporal blending providing a more significant improvement. Ultimately, combining both semantic conditioning and temporal blending results in the best FVD score, highlighting their complementary benefits and demonstrating the importance of these components in achieving the best translation quality; see Fig.~\ref{fig:benefits_sem_tb}.

\paragraph{Effects of decay factor $\omega$} The magnitude of directional guidance for denoising trajectory is controlled by the decay factor $\omega$ (see Algo.~\ref{algo:temporal_blend}). Setting $\omega = 0$ is equivalent to completely disabling the temporal blending, resulting in a high FVD score due to the inter-frame inconsistencies. Conversely, setting $\omega = 1$ applies strict guidance at every denoising step, significantly improving FVD; see Fig.~\ref{fig:decay_effects}. However, the best FVD is achieved by gradually reducing the guidance magnitude over time, balancing the translation quality and inter-frame smoothness; see Fig.~\ref{fig:decay_effects}.

\paragraph{Why segmentation logits?} We propose using segmentation logits instead of masks for semantic conditioning; see Fig.~\ref{fig:i2v-pdm}. This was motivated by the hypothesis that the foundational segmentation model~\cite{dinov2}, typically trained on visible images, could introduce prediction errors when directly applied to infrared images due to the domain gap between two modalities. To address this, we utilised segmentation logits as soft predictions, mitigating potential errors associated with hard predictions like segmentation masks. As a result, using segmentation logits consistently offers the best FVD across all training iterations, outperforming the use of masks; see Fig.~\ref{fig:masks_vs_logits}.

\subsection{Day~$\rightarrow$~night object detection} \label{sec:exp_obj_det}
Given a training dataset containing daytime visible images and groundtruth bounding boxes (source domain), the goal is to train an object detector that can self-adapt to perform effectively at night (target domain). If the adaptation must happen without access to the source domain due to privacy constraints, this is known as test-time adaptation~\cite{dua}. To address this challenge, we use \ourmethod to translate nighttime infrared (or visible) images to their daytime visible counterparts. The object detector, pre-trained on the source domain, can then be directly applied to detect objects. Here, we employ YOLOv5 as the object detector.

\paragraph{Dataset} We used M3FD dataset~\cite{m3fd} containing 4,200 infrared-visible image pairs. These were extracted from videos captured across various times and scenes, including a university campus, a tourist resort, and city locations. We split the dataset into daytime and nighttime sets, and then selected images with cars. All images were resized and cropped to 256$\times$256, resulting in 1,945 daytime pairs and 608 nighttime pairs for training and testing. \ourmethod is trained on the daytime visible-infrared pairs and YOLOv5 is trained on the visible daytime images to detect cars. Then, we tested \ourmethod and YOLOv5 on the nighttime set.

\paragraph{Baselines} In addition to comparing our method to other translation approaches presented in Sec.~\ref{sec:exp:i2v}, we also selected two test-time adaptation methods BN Stats~\cite{bnstats} and DUA~\cite{dua} for comparison.

\paragraph{Metrics} We reported the average precision at a 50\% threshold (\textbf{AP}$_{50}$) and across multiple thresholds from 50\% to 95\% with 5\% increments (\textbf{AP}$_{50:95}$).

\begin{table}
    \centering
    \small
    \begin{tabular}{llcc}
        \toprule
          Strategy & Method & AP$_{50}$$\uparrow$  & AP$_{50:95}$$\uparrow$ \\
          \midrule
          \textbf{No} &Infrared & 43.8 & 21.7 \\
          \textbf{adaptation} & Visible & 43.3 & 27.7 \\
          \midrule
          & BN Stats (Infrared)  &  44.5 & 22.1\\
          \textbf{Test-time} & BN Stats (Visible) & 44.0 & 27.8\\
          \textbf{adaptation} &  DUA (Infrared) & 46.4 & 23.1 \\
          &  DUA (Visible) & 44.1 & 27.8 \\
            
          \midrule
         & Recycle-GAN &  36.4 & 18.6 \\
         &UnsuperRecycle & 14.6 & 4.2 \\
         \textbf{Infrared-to}&I2V-GAN & 44.5 & 21.3 \\
        \textbf{visible} & ROMA & 57.2 & 34.7 \\
         \textbf{translation} & T2V-DDPM & 56.7 & 33.0\\
         &PDIR & 62.9 & 37.5 \\
         &\textbf{\ourmethod (Ours)} & \textbf{69.0} & \textbf{42.3}\\
         \bottomrule
    \end{tabular}
    \caption{Comparison of day~$\rightarrow$~night object detection performance on M3FD dataset.} 
    \label{tab:obj_det}
\end{table}

\paragraph{Results} In general, our method significantly outperforms existing approaches; see Tab.~\ref{tab:obj_det}. Specifically, BN Stats and DUA yielded a marginal gain (1-2\%) when applied to either infrared or visible modality. Also, Recycle-GAN and UnsuperRecycle decreased the accuracy, whereas I2V-GAN showed a slight improvement of less than 1\%. However, ROMA, T2V-DDPM, and PDIR significantly enhanced the performance by 10\%-20\% in  AP$_{50}$ and 10\%-15\% in AP$_{50:95}$. Compared to these methods, \ourmethod achieved the best performance, improving the AP$_{50}$ and AP$_{50:95}$ by about 25\% and 20\% respectively; see also Fig.~\ref{fig:quantitative_quality} for the qualitative results of I2V translation. More qualitative I2V translation results and object detection are respectively shown in Secs.~\ref{supp:sec:i2v_more_results} and~\ref{supp:sec:obj_det} in the supplementary material.

\section{Conclusion}
This paper presents \ourmethod for I2V video translation, comprising two key components: (i) semantic conditioning to preserve semantic structure in generated visible images, and (ii) temporal blending to ensure smooth inter-frame transitions. The experiment results show that \ourmethod outperforms state-of-the-art methods in I2V video translation and day~$\rightarrow$~night object detection.

\section*{Acknowledgment}
This project is supported by the Australian Research Council (ARC) project number LP200200881, Safran Electronics
and Defense Australasia, and Safran Electronics and Defense.

\section*{Supplementary Material}
\begin{alphasection}

\section{Infrared-to-visible translation: more results}
\label{supp:sec:i2v_more_results}
Figs.~\ref{fig:qualitative_kaist_more} and~\ref{fig:qualitative_m3fd_more} respectively show more qualitative results on KAIST and M3FD datasets.

\begin{table}[b]
    \centering
    \begin{tabular}{lcccc}
        \toprule
          & \textbf{Realism}$\uparrow$  & \textbf{Smoothness}$\uparrow$  \\
          \midrule
          Recycle-GAN & 2.4 $\pm$ 1.5 & 4.6 $\pm$ 1.7 \\
         UnsuperRecycle & 2.0 $\pm$ 1.2 & 2.1 $\pm$ 1.2 \\
         I2V-GAN & 4.9 $\pm$ 1.8 & 6.1 $\pm$ 2.3\\
         ROMA & 6.4 $\pm$ 1.6 & 6.9 $\pm$ 1.5 \\
         T2V-DDPM & 5.9 $\pm$ 1.5 & 5.9 $\pm$ 1.5 \\
         PDIR & 4.3 $\pm$ 2.2 & 2.7 $\pm$ 1.3\\
         \textbf{\ourmethod (Ours)} & \textbf{8.4	$\pm$ 1.2} & \textbf{8.6 $\pm$ 1.6} \\
         \bottomrule
    \end{tabular}
    \caption{User Study: Infrared-to-Visible Video Translation on KAIST Dataset. Using a $10$-point scoring system (where $10$ is the highest), participants are requested to rate each video's realism and smoothness, where higher scores indicate better performance in each metric.}
    \label{tab:user_study}
\end{table}

\paragraph{User study} Following~\cite{recyclegan, i2vgan, wang2018video, wang2019few}, we conducted a user study on generated visible videos from the KAIST dataset.
In particular, participants were shown
\begin{itemize}
    \item Input infrared videos
    \item Actual visible videos
    \item Generated visible videos (randomly arranged, without revealing the generation method)
\end{itemize}
We then defined two metrics to evaluate the results
\begin{itemize}
    \item \textbf{Realism} measures how closely the generated visible video resembles the actual visible video in terms of visual fidelity and authenticity.
    \item \textbf{Smoothness:}  evaluates the temporal continuity of the generated video, specifically the fluent transitions between consecutive frames.
\end{itemize}
We asked $18$ participants to evaluate the realism and smoothness of the generated videos, using a scoring system ranging from $1$ to $10$. A higher score indicates better performance in each metric.

The result is shown in Tab.~\ref{tab:user_study}. Overall, participants consistently rated \ourmethod highest, with mean scores of $8.4$ for realism and $8.6$ for smoothness. Interestingly, despite PDIR's competitive FID scores, participants rated its realism with a low score ($4.3$). This discrepancy highlights the significant impact of  temporal inconsistency on human perception, showing that visual fidelity evaluations are considerably influenced by temporal coherence. This finding emphasises the importance of ensuring temporal consistency in video generation.
\begin{figure*}
    \centering
    \includegraphics[width=1.0\linewidth]{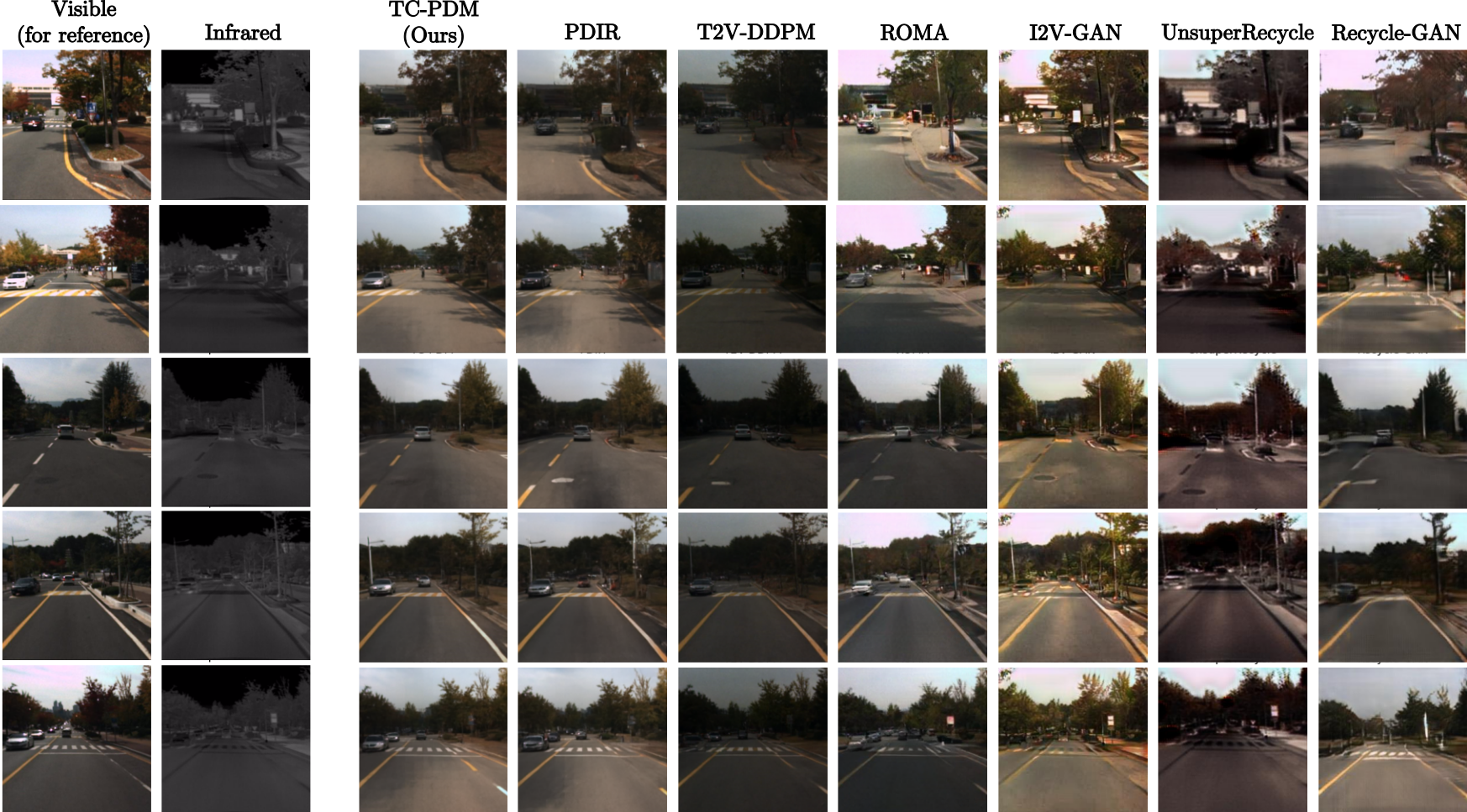}
    \caption{Qualitative comparison of different translation methods on the KAIST dataset.}
    \label{fig:qualitative_kaist_more}
\end{figure*}

\begin{figure*}
    \centering
    \includegraphics[width=1.0\linewidth]{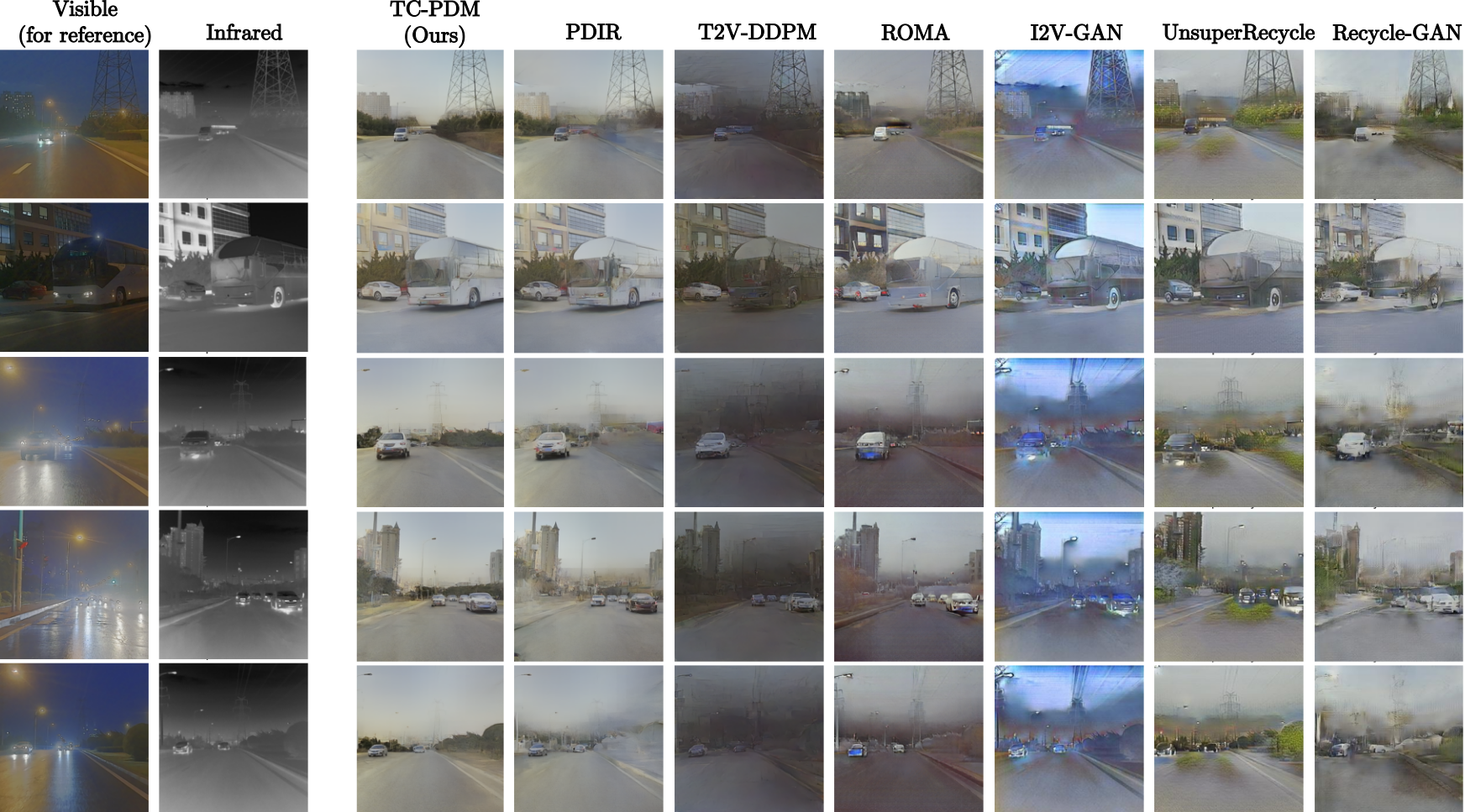}
    \caption{Qualitative comparison of different translation methods on the M3FD dataset.}
    \label{fig:qualitative_m3fd_more}
\end{figure*}

\section{Object detection: qualitative results}
\label{supp:sec:obj_det}

Fig.~\ref{fig:quant_obj_det} shows the qualitative results of day~$\rightarrow$~night object detection.
\begin{figure*}[t]
    \centering
    \includegraphics[width=1.0\linewidth]{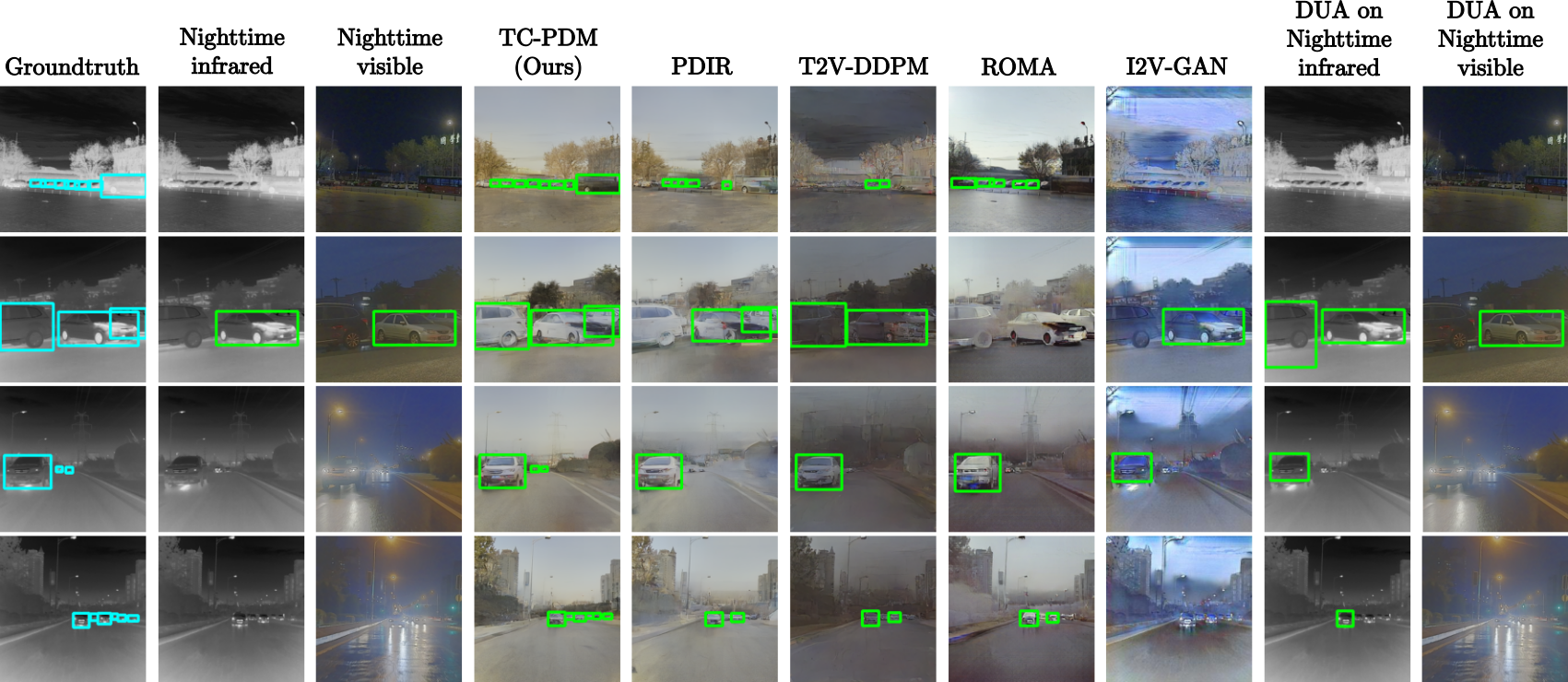}
    \caption{Qualitative comparison of day~$\rightarrow$~night object detection performance on M3FD dataset. \ourmethod enabled YOLOv5 detector to accurately detect small objects.}
    \label{fig:quant_obj_det}
\end{figure*}

\end{alphasection}

{
    \small
    \bibliographystyle{ieeenat_fullname}
    \bibliography{main}
}


\end{document}